\documentclass{article}
\renewcommand{\subparagraph}{\paragraph}
\usepackage{include/nips13submit_e}
\usepackage{times}
\usepackage{amsmath, amsthm, amssymb}
\usepackage{pstricks}
\usepackage{pst-tree}
\usepackage{bm}
\usepackage{graphicx}
\usepackage{epsfig}
\usepackage{caption}
\usepackage{subcaption}
\usepackage{floatrow}
\usepackage{amsmath}
\usepackage{amsfonts}
\usepackage{color}
\usepackage{array}
\usepackage{colortbl}
\usepackage{framed}
\usepackage{url}
\usepackage{booktabs}
\usepackage{multirow}
\usepackage{dsfont}

\usepackage{multicol}
\usepackage{lscape}
\usepackage{relsize}
\usepackage{rotating}
\usepackage{tikz}
\usetikzlibrary{calc}
\usepackage{titlesec}

\usepackage{enumitem}
\setlist[itemize]{noitemsep, topsep=0pt} % Makes els 1 and 2 fit more nicely
\usepackage{nicefrac}
\newcommand{\vect}[1]{\underline{\smash{#1}}}
\renewcommand{\v}[1]{\vect{#1}}
\newcommand{\reals}{\mathds{R}}
\newcommand{\sX}{\mathcal{X}}

\newcommand{\br}{}%{^{\text{\textnormal{ r}}}}

\newcommand{\cut}[1]{}
\newcommand{\hide}[1]{}

\DeclareFontFamily{OT1}{pzc}{}
\DeclareFontShape{OT1}{pzc}{m}{it}{<-> s * [1.200] pzcmi7t}{}
\DeclareMathAlphabet{\mathscr}{OT1}{pzc}{m}{it}

\newcommand\transpose{{\textrm{\tiny{\sf{T}}}}}
\newcommand{\note}[1]{}

\usepackage{amsmath, amsthm, amssymb}
\newcommand{\embeddingletter}{g}
\newcommand{\bo}{{\sc bo}}
\newcommand{\gp}{{\sc gp}}
\newcommand{\agp}{Arc \gp}

\nipsfinalcopy

\begin{document}

\title{Raiders of the Lost Architecture:\\Kernels for Bayesian Optimization in Conditional Parameter Spaces}

\author{
Kevin Swersky \\
University of Toronto \\
{\small \texttt{kswersky@cs.toronto.edu}} \\
\And
David Duvenaud \\
University of Cambridge \\
{\small \texttt{dkd23@cam.ac.uk}} \\
\And
Jasper Snoek\\
Harvard University \\
{\small \texttt{jsnoek@seas.harvard.edu}} \\
\AND
Frank Hutter  \\
Freiburg University \\
{\small {\tt fh@informatik.uni-freiburg.de}} \\
\And
Michael A. Osborne \\
University of Oxford \\
{\small{\tt mosb@robots.ox.ac.uk}} \\
}

\maketitle
\begin{abstract}
In practical Bayesian optimization, we must often search over structures with differing numbers of parameters.  For instance, we may wish to search over neural network architectures with an unknown number of layers.  To relate performance data gathered for different architectures, we define a new kernel for conditional parameter spaces that explicitly includes information about which parameters are relevant in a given structure. We show that this kernel improves model quality and Bayesian optimization results over several simpler baseline kernels.
\end{abstract}

%\note{FH: I left notes throughout using this mechanism. These notes stand out ugly on purpose -- so that they can't be overlooked easily. Once a note is dealt with, please remove it or comment it out in the source. For checking e.g. length, all notes can also be disabled at once by commenting out a single line towards the top of the source.}

%\note{FH: changed title to include ``Bayesian Optimization in'' since its the BayesOpt workshop. I mildly prefer that but it's longer, so I'm also happy if someone wants to undo hte change.}
%\note{MAO: I agree with title change, at least for this version of the paper. }

%%%%%%%%%%%%%%%%%%%%%%%%%%%%%%%%%%%%%%%%%%%%%%%%%%%%%%%%%%%%%%%%%%%%%%%%%%%%%%%%%%%%%%%%%%%%%%%%%
\section{Introduction}
%%%%%%%%%%%%%%%%%%%%%%%%%%%%%%%%%%%%%%%%%%%%%%%%%%%%%%%%%%%%%%%%%%%%%%%%%%%%%%%%%%%%%%%%%%%%%%%%%
%\vspace{-0.05in} 

%FH: substituted this nice description with a more concise version - we don't need to tell people at BayesOpt in detail what Bayesian optimization is.

Bayesian optimization (\bo) is an efficient approach for solving blackbox optimization problems of the form $\arg\min_{x \in X} f(x)$ (see~\cite{Brochu2010} for a detailed overview), where $f$ is expensive to evaluate. 
It employs a prior distribution $p(f)$ over functions that is updated as new information on $f$ becomes available.
The most common choice of prior distribution are Gaussian processes (\gp s~\cite{rasmussen38gaussian}), as they are powerful and flexible models for which the marginal and conditional distributions can be computed efficiently.\footnote{There are prominent exceptions to this rule, though. In particular, tree-based models, such as random forests, can be a better choice if there are many data points (and \gp s thus become computationally inefficient), if the input dimensionality is high, if the noise is not normally distributed, or if there are non-stationarities~\cite{TadGraPol11,HutHooLey11,bergstra2011algorithms}.}
However, some problem domains remain challenging to model well with \gp s, and the efficiency and effectiveness of Bayesian optimization suffers as a result. In this paper, we tackle the common problem of input dimensions that are only relevant if other inputs take certain values~\cite{Hut09:phd,bergstra2011algorithms}. This is a general problem in algorithm configuration~\cite{Hut09:phd} that occurs in many machine learning contexts, such as, e.g., in deep neural networks~\cite{HinOsiTeh06}; flexible computer vision architectures~\cite{BerYamCox13}; and the combined selection and hyperparameter optimization of machine learning algorithms~\cite{ThoEtAl13}. We detail the case of deep neural networks below.

Bayesian optimization has recently been applied successfully to deep neural networks~\cite{snoek-etal-2012b, bergstra2011algorithms} to optimize high level model parameters and optimization parameters, which we will refer to collectively as \emph{hyperparameters}.  Deep neural networks represent the state-of-the-art on multiple machine learning benchmarks such as object recognition~\cite{krizhevsky-2012}, speech recognition~\cite{deepSpeechReviewSPM2012}, natural language processing~\cite{mikolov2010recurrent} and more.
They are multi-layered models by definition, and each layer is typically parameterized by a unique set of hyperparameters, such as regularization parameters and the layer capacity or number of hidden units.  Thus adding additional layers introduces additional hyperparameters to be optimized.  The result is a complex hierarchical conditional parameter space, which is difficult to search over.  Historically, practitioners have simply built a separate model for each type of architecture or used non-\gp~models~\cite{bergstra2011algorithms}, or assumed a fixed architecture~\cite{snoek-etal-2012b}.  If there is any relation between networks with different architectures, separately modelling each is wasteful. 

\gp s with standard kernels fail to model the performance of architectures with such conditional hyperparameters. To remedy this, 
the contribution of this paper is the introduction of a kernel that allows observed information to be shared across architectures when this is appropriate. We demonstrate the effectiveness of this kernel on a \gp{} regression task and a Bayesian optimization task using a feed-forward classification neural network.

\section{A Kernel for Conditional Parameter Spaces}
%\vspace{-0.05in} 

\gp{}s employ a positive-definite kernel function $k: \sX \times \sX \rightarrow \mathbb{R}$ to model the covariance between function values. Typical \gp{} models cannot, however, model the covariance between function values whose inputs have different (possibly overlapping) sets of relevant variables.

In this section, we construct a kernel between points in a space that may have dimensions which are irrelevant under known conditions (further details are available in \cite{arxiv_hierarchical_kernel}). As an explicit example, we consider a deep neural network: if we set the network depth to 2 we know that the 3rd layer's hyperparameters do not have any effect (as there is no 3rd layer).

Formally, we aim to do inference about some function $f$ with domain 
%(input space)
 $\sX$. $\sX = \prod_{i=1}^D \sX_i$ is a $D$-dimensional input space, where each individual dimension is bounded real, that is, $\sX_i = [l_i, u_i] \subset \reals$ (with lower and upper bounds $l_i$ and $u_i$, respectively). We define functions $\delta_i\colon \sX\to \{\text{true}, \text{false}\}$, for $i \in \{1,\,\ldots,\,D\}$. $\delta_i(\v{x})$ stipulates the relevance of the $i$th feature $x_i$ to 
 %inference about
  $f(\v{x})$.

\subsection{The problem}
%\vspace{-0.05in}

As an example, imagine trying to model the performance of a neural network having either one or two hidden layers, with respect to the regularization parameters for each layer, $x_1$ and $x_2$.  If $y$ represents the performance of a one layer-net with regularization parameters $x_1$ and $x_2$, then the value $x_2$ doesn't matter, since there is no second layer to the network. Below, we'll write an input triple as $(x_1, \delta_2(\v{x}), x_2)$ and assume that $\delta_1(\v{x}) = \text{true}$; that is, the regularization parameter for the first layer is always relevant. 

In this setting, we want a kernel $k$ to be dependent on which parameters are relevant, and the values of relevant parameters for both points. For example, consider first-layer parameters $x_1$ and $x_1'$:
\begin{itemize}[leftmargin=0.8cm]
\item If we are comparing two points for which the same parameters are relevant, the value of any unused parameters shouldn't matter,  
\begin{equation}
 k\bigl((x_1, \textnormal{false}, x_2), (x_1', \textnormal{false}, x_2') \bigr)
= k\bigl((x_1, \textnormal{false}, x_2''), (x_1', \textnormal{false}, x_2''')\bigr),\ 
\forall x_2, x_2', x_2'', x_2''';
\end{equation}
\item The covariance between a point using both parameters and a point using only one should again only depend on their shared parameters,
\begin{equation}
 k\bigl((x_1, \textnormal{false}, x_2), (x_1', \textnormal{true}, x_2') \bigr)
= k\bigl((x_1, \textnormal{false}, x_2''), (x_1', \textnormal{true}, x_2''')\bigr),\ 
\forall x_2, x_2', x_2'', x_2'''.
\end{equation}
\end{itemize}

Put another way, in the absence of any other information, this specification encodes our prior ignorance about the irrelevant (missing) parameters while still allowing us to model correlations between relevant parameters.
%Elaborating on the first consideration:
%
%\begin{itemize}[leftmargin=0.8cm]
%\item The covariance between two points with identical values for their jointly relevant parameters should only depend on whether their remaining parameters are relevant,
%\begin{align}
% k\bigl((x_1, \textnormal{false}, x_2), (x_1, \textnormal{false}, x_2') \bigr)
%& = k_{\text{FF}},\ \forall x_2, x_2'\\
% k\bigl((x_1, \textnormal{false}, x_2), (x_1, \textnormal{true}, x_2') \bigr)
%& = k_{\text{FT}},\ \forall x_2, x_2'.
%\end{align}
%\end{itemize}
%We usually additionally want $k_{\text{FF}}>k_{\text{FT}}$, expressing the fact that points that have identical relevance $\delta_2({\v{x}})$ are more similar than points that differ in relevance $\delta_2({\v{x}})$.

\subsection{Cylindrical Embedding}
%\vspace{-0.05in} 

We can build a kernel with these properties for each possibly irrelevant input dimension $i$ by embedding our points into a Euclidean space.  Specifically, the embedding we use is
%
%To emphasize that we're in the real case, we explicitly denote the pseudometric as $d\br_i$ and the (pseudo-)isometry from $(\sX, d_i)$ to $\reals^2,d_\text{E}$ 
%as $f\br_i$. For the definitions, recall that $\delta_i(\v{x})$ is true iff dimension $i$ is relevant given the instantiation of $i$'s ancestors in $\v{x}$.
%
%
\begin{equation}
\embeddingletter_i\br(\v{x}) = \left\{\begin{array}{ll}
[0,0]^\transpose & \textrm{ if } \delta_i(\v{x}) = \textrm{ false }\\
\omega_i [\sin{\pi\rho_i\frac{x_i}{u_i-l_i}}, \cos{\pi\rho_i\frac{x_i}{u_i-l_i}}]^\transpose & \textrm{ otherwise.}\end{array}\right.
\label{eq:embedding}
\end{equation}
Where $\omega_i \in \mathbb{R}^+$ and $\rho_i \in [0,1]$.
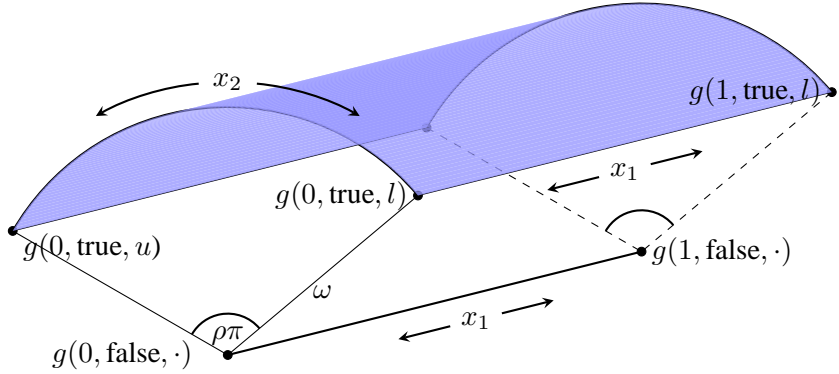
\begin{figure}
% A simple figure to illustrate Mike and Frank's embedding
% Sept 2013

\newcommand{\scaleamount}{1.1}
\newcommand{\anglemin}{40}
\newcommand{\angleminplusone}{41}
\newcommand{\anglemax}{150}
\newcommand{\angleone}{70}
\newcommand{\angletwo}{120}
\newcommand{\bigradius}{3}
\newcommand{\biggerradius}{3.3}
\newcommand{\smallradius}{0.5}
\newcommand{\xtwolabelangle}{90}
\newcommand{\belowamount}{0.2cm}
\newcommand{\belowamounttwo}{0.11cm}
\newcommand{\myell}{l}
\newcommand{\embedding}{g}

%\begin{center}
%\framebox{
\begin{tikzpicture}
[trans/.style={thick,->,shorten >=2pt,shorten <=2pt,>=stealth},scale=\scaleamount, every node/.style={transform shape}]

\tikzset{state/.style={circle,draw=black, very thick,minimum size=4em}}

\def\centerarc[#1] (#2) (#3:#4:#5)% [draw options] (center) (initial angle:final angle:radius)
{ \draw[#1] (#2) ++(#3:#5) arc (#3:#4:#5);
}
	\coordinate (left) at ({\bigradius*cos(\anglemax)}, {\bigradius*sin(\anglemax)});
	\coordinate (right) at ({\bigradius*cos(\anglemin)}, {\bigradius*sin(\anglemin)});

	\coordinate (O1) at (-2, 0);
	\coordinate (O2) at (3, 1.25);
	
	\coordinate (left1) at ($ (left) + (O1) $);
	\coordinate (right1) at ($ (right) + (O1) $);
	\coordinate (left2) at ($ (left) + (O2) $);
	\coordinate (right2) at ($ (right) + (O2) $);
	
	% First arc
	% ==================
	
	% Draw arcs
	\centerarc[thick] (O1) (\anglemin:\anglemax:\bigradius)
	\centerarc[thick] (O1) (\anglemin:\anglemax:\smallradius)

	\draw[fill] (left1) circle (1.5pt);
	\draw (left1) node[below = \belowamount, right] {$\embedding(0, \textnormal{true}, u$)};
	
	\draw[fill] (right1) circle (1.5pt);
	\draw (right1) node[below = \belowamount] (r1below) {}
	               node[left] {$\embedding(0, \textnormal{true}, \myell)$};

	\draw[fill] (O1) circle (1.5pt);
	\draw (O1) node[below = \belowamount] (o1below) {}
	           node[left = 0.3cm] {$\embedding(0, \textnormal{false}, \cdot)$};

	\draw (left1) -- (O1) node[above] {$\rho \pi$};
	\draw (right1) -- (O1) node[midway, below] {$\omega$};

	% Second arc
	% ================================
	\centerarc[thick] (O2) (\anglemin:\anglemax:\bigradius)
	\centerarc[thick] (O2) (\anglemin:\anglemax:\smallradius)

	\draw[fill] (left2) circle (1.5pt);
	
	\draw[fill] (right2) circle (1.5pt);
	\draw (right2) node[below = \belowamount] (r2below) {};

	\draw[fill] (O2) circle (1.5pt);
	\draw (O2) node[below = \belowamount] (o2below) {}
                   node[right] {$\embedding(1, \textnormal{false}, \cdot)$};

	\draw[dashed] (left2) -- (O2);
	\draw[dashed] (right2) -- (O2);

	% Connect the arcs
	\draw (left1) -- (left2);
	\draw (right1) -- (right2);
	\draw[thick] (O1) -- (O2);
	
	% Draw surface
	\foreach \i in {\angleminplusone, ..., \anglemax}
	{
		\coordinate (arca) at ({\bigradius*cos(\i)}, {\bigradius*sin(\i)});
		\coordinate (arcb) at ({\bigradius*cos(\i - 1)}, {\bigradius*sin(\i - 1)});
		\coordinate (arca1) at ($ (arca) + (O1) $);
		\coordinate (arca2) at ($ (arca) + (O2) $);
		\coordinate (arcb1) at ($ (arcb) + (O1) $);
		\coordinate (arcb2) at ($ (arcb) + (O2) $);
%		\draw[green] (arca1) -- (arca2);
		\fill[fill=blue!40,fill opacity=0.8](arca1)--(arca2)--(arcb2)--(arcb1)--cycle;
	}

	\draw (right2) node[left] {$\embedding(1, \textnormal{true}, \myell)$};

	% Draw arrows showing in which direction x1 varies.
	\coordinate(x1half) at ($ (right1)!0.5!(right2) $);
	\draw (x1half) node[below = \belowamounttwo] (x1halfbelow) {$x_1$};
	\draw[trans] ($ (r1below)!0.45!(r2below) $) -- ($ (r1below)!0.3!(r2below) $);
	\draw[trans] ($ (r1below)!0.55!(r2below) $) -- ($ (r1below)!0.7!(r2below) $);

	\coordinate(x2half) at ($ (O1)!0.6!(O2) $);
	\draw (x2half) node[below = \belowamounttwo] (x2halfbelow) {$x_1$};
	\draw[trans] ($ (o1below)!0.55!(o2below) $) -- ($ (o1below)!0.4!(o2below) $);
	\draw[trans] ($ (o1below)!0.65!(o2below) $) -- ($ (o1below)!0.8!(o2below) $);

	% Draw arrows showing in which direction x2 varies.
	\coordinate (arclabel) at ($ (O1) + ({\biggerradius*cos(\xtwolabelangle)}, {\biggerradius*sin(\xtwolabelangle)}) $);
	\draw (arclabel) node {$x_2$};
	\centerarc[trans] (O1) ( \xtwolabelangle + 5  :  \xtwolabelangle + 30  :\biggerradius)
	\centerarc[trans] (O1) ( \xtwolabelangle - 5  :  \xtwolabelangle - 30  :\biggerradius)

%	\coordinate (x1) at ({\bigradius*cos(\angletwo)}, {\bigradius*sin(\angletwo)});
%	\coordinate (x2) at ({\bigradius*cos(\angleone)}, {\bigradius*sin(\angleone)});
%	\draw[fill] (x1) circle (1.5pt);
%	\draw (x1) node[above, left] {$f(x_1, \textnormal{true})$};
%	\draw[fill] (x2) circle (1.5pt);
%	\draw (x2) node[above, right] {$f(x_2, \textnormal{true})$};
\end{tikzpicture}
%}
%\end{center}
%	\floatbox[{\capbeside\thisfloatsetup{capbesideposition={right,top}}}]{figure}[\FBwidth]
\caption{A demonstration of the embedding giving rise to the pseudo-metric.  All points for which $\delta_2(x) =$ false are mapped onto a line varying only along $x_1$.  Points for which $\delta_2(x) =$ true are mapped to the surface of a semicylinder, depending on both $x_1$ and $x_2$.  This embedding gives a constant distance between pairs of points which have differing values of $\delta$ but the same values of $x_1$.}
	{\hspace{-1cm}\label{fig:cylinder}}
	%\caption{}
%  The parameter $\rho$ determines how much distance there is along the arc.
%\vspace{-0.3cm}
\end{figure}

Figure \ref{fig:cylinder} shows a visualization of the embedding of points $(x_1, \delta_2(\v{x}), x_2)$ into $\reals^3$. 
In this space, we have the Euclidean distance,
\begin{equation}
d\br_i(\v{x}, \v{x}') = ||\embeddingletter_i\br(\v{x})-\embeddingletter_i\br(\v{x}')||_2 =\left\{\begin{array}{ll}
0 & \textrm{ if } \delta_i(\v{x}) = \delta_i(\v{x}') = \textrm{false}\\
\omega_i & \textrm{ if } \delta_i(\v{x}) \neq \delta_i(\v{x}')\\
\omega_i \sqrt{2} \sqrt{1 - \cos(\pi\rho_i \frac{x_i-x_i'}{u_i-l_i})} & \textrm{ if } \delta_i(\v{x}) = \delta_i(\v{x}') = \textrm{true}. \end{array}\right.
\label{eq:distance}
\end{equation}

We can use this to define a covariance over our original space. In particular, we consider the class of covariances that are functions only of the Euclidean distance $\Delta$ between points. There are many examples of such covariances. Popular examples are the exponentiated quadratic, for which $\kappa(\Delta) = \sigma^2 \exp(-\frac{1}{2} \Delta^2)$, or the rational quadratic, for which $\kappa(\Delta) = \sigma^2 (1+\frac{1}{2\alpha} \Delta^2)^{-\alpha}$. We can simply take \eqref{eq:distance} in the place of $\Delta$, returning a valid covariance that satisfies all desiderata above.

Explicitly, note that as desired, if $i$ is irrelevant for both $\v{x}$ and $\v{x}'$, $d\br_i$ specifies that $g(\v{x})$ and $g(\v{x}')$ should not differ owing to differences between $x_i$ and $x_i'$. Secondly, if $i$ is relevant for both $\v{x}$ and $\v{x}'$, the difference between $f(\v{x})$ and $f(\v{x}')$ due to $x_i$ and $x_i'$ increases monotonically with increasing $\left|x_i-x_i'\right|$. The parameter $\rho_i$ controls whether differing in the relevance of $i$ contributes more or less to the distance than differing in the value of $x_i$, should $i$ be relevant. 
%
%
%\note{FH: the following 3 sentences are details that could be dropped if we're in need of space.}
%If $\rho = \nicefrac{1}{3}$, and if $i$ is irrelevant for exactly one of $\v{x}$ and $\v{x}'$, $f(\v{x})$ and $f(\v{x}')$ are as different as is possible due to dimension $i$; that is, $f(\v{x})$ and $f(\v{x}')$ are exactly as different in that case as if $x_i=l_i$ and $x_i'=u_i$. For $\rho>\nicefrac{1}{3}$, $i$ being relevant for both $\v{x}$ and $\v{x}'$ means that $f(\v{x})$ and $f(\v{x}')$ could potentially be more different than if %$i$ was relevant in only one of them. For $\rho<\nicefrac{1}{3}$, the converse is true. 
%
%
Hyperparameter $\omega_i$ defines a length scale for the $i$th feature. 

Note that so far we only have defined a kernel for dimension $i$. To obtain a kernel for the entire $D$-dimensional input space, we simply embed each dimension in $\mathbb{R}^2$ using Equation~(\ref{eq:embedding}) and then use the embedded input space of size $2D$ within any kernel that is defined in terms of Euclidean distance.
We dub this new kernel the \emph{arc kernel}. Its parameters, $\omega_i$ and $\rho_i$ for each dimension, can be optimized using the \gp{} marginal likelihood, or integrated out using Markov chain Monte Carlo.

\section{Experiments}
%\vspace{-0.05in} 

We now show that the arc kernel yields better results than other alternatives. 
% Bayesian optimization requires building a surrogate model of the function being optimized, and better models can be expected to lead to more efficient optimization.  However, because of the many interacting components of BO, optimizer performance might not correspond directly to the quality of the model.  Thus, w
We perform two types of experiments: first, we study model quality in isolation in a regression task; second, we study the effect of the arc kernel on \bo{} performance. All \gp{} models use a Mat\'{e}rn $\nicefrac{5}{2}$ kernel.

\paragraph{Data.} We use two different datasets, both of which are common in the deep learning literature. The first is the canonical MNIST digits dataset~\cite{lecun-1998a} where the task is to classify handwritten digits. The second is the CIFAR-10 object recognition dataset~\cite{Krizhevsky-2009a}. We pre-processed CIFAR-10 by extracting features according to the pipeline given in~\cite{coates2010analysis}.

\subsection{Model Quality Experiments}   

\paragraph{Models.}

Our first experiments concern the quality of the regression models used to form the response surface for Bayesian optimization. We generated data by performing 10 independent runs of Bayesian optimization on MNIST and then treat this as a regression problem. We compare the \gp{} with arc kernel (\agp) to several baselines: the first baseline is a simple linear regression model, the second is a \gp{}  where irrelevant dimensions are simply filled in randomly for each input. We also compare to the case where each architecture uses its own separate \gp{}, as in~\cite{bergstra2011algorithms}. The results are averaged over $10$-fold train/test splits. Kernel parameters were inferred using slice sampling \cite{Murray-Adams-2010a}. As the errors lie between $0$ and $1$ with many distributed toward the lower end, it can be beneficial to take the log of the outputs before modelling them with a \gp{}. We experiment with both the original and transformed outputs.

%Separate Linear and Seperate {\sc gp} build a separate model for each neural net architecture, as in \cite{bergstra2011algorithms}.  The hierarchical {\sc gp} model combines all data together using the conditional kernel.
%
%In the case where all data comes from the same architecture, the hierarchical {\sc gp} model makes slightly different assumptions than a standard {\sc gp}, because it places the data on a semi-circle.  To check whether this alternate assumption affects performance, we also include a Separate-Hierarchical {\sc gp} model.
%
%We also compare against a simpler embedding method, or ``Poor Man's Embedding''.  In this procedure, we combine data from all models into a single model, replacing any unused parameters with a heuristic constant, set to $-1$ in these experiments.

%\paragraph{Data.} 951 datapoints were generated by performing Bayesian optimization on MNIST.  Results are averaged over 10-fold train/test splits.
%\note{FH: very briefly describe the number of data points, where they come from, and how they were split to obtain error bars (I assume cross-validation).}

\begin{table}[h!]
\caption{{\small Normalized Mean Squared Error on MNIST Bayesian optimization data\label{tab:nn_error}}}
\label{tbl:nn_nmse}
% --- Automatically generated by resultsToLatex4.m ---
% Exported at 20-Oct-2013 19:39:08
\begin{center}
\begin{tabular}{l | r r}
Method & \rotatebox{0}{ Original data   }  & \rotatebox{0}{ Log outputs }  \\ \hline
Separate Linear & $0.812 \pm 0.045$ & $0.737 \pm 0.049$ \\
Separate \gp{} & $0.546 \pm  0.038$ & $0.446 \pm 0.041$ \\
Separate \agp{} & $0.535 \pm 0.030$ & $0.440 \pm 0.031$ \\
Linear & $0.876 \pm 0.043$ & $0.834 \pm 0.047$ \\
\gp{} & $0.481 \pm 0.031$ & $0.401 \pm 0.028$ \\
\agp{} & $\mathbf{0.421 \pm  0.033}$ & $\mathbf{0.335 \pm 0.028}$
\end{tabular}
\end{center}
% End automatically generated LaTeX

%\input{tables/mcmc-table-transposed.tex}
%\caption{Regression errors for a GP with the arc kernel compared to baselines.}
%\vspace{-0.3cm}
\end{table}

\paragraph{Results.}
Table \ref{tbl:nn_nmse} shows that a \gp{} using the arc kernel performs favourably to a \gp{} that ignores the relevance information of each point. The ``separate'' categories apply a different model to each layer and therefore do not take advantage of dependencies between layers. Interestingly, the separate Arc \gp{}, which is effectively just a standard \gp{} with additional embedding, performs comparably to a standard \gp{}, suggesting that the embedding doesn't limit the expressiveness of the model.

\subsection{Bayesian Optimization Experiments}  
%\vspace{-0.05in} 
In this experiment, we test the ability of Bayesian optimization to tune the hyperparameters of each layer of a deep neural network. We allow the neural networks for these problems to use up to $5$ hidden layers (or no hidden layer). We optimize over learning rates, L2 weight constraints, dropout rates~\cite{hinton2012improving}, and the number of hidden units per layer leading to a total of up to $23$ hyperparameters and $6$ architectures. On MNIST, most effort is spent improving the error by a fraction of a percent, therefore we optimize this dataset using the log-classification error. For CIFAR-10, we use classification error as the objective. We use the Deepnet\footnote{https://github.com/nitishsrivastava/deepnet} package, and each function evaluation took approximately $1000$ to $2000$ seconds to run on NVIDIA GTX Titan GPUs. Note that when a network of depth $n$ is tested, all hyperparameters from layers $n+1$ onward are deemed irrelevant.
\paragraph{Experimental Setup.}
For Bayesian optimization, we follow the methodology of~\cite{snoek-etal-2012b}, using slice sampling and the expected improvement heuristic. In this methodology, the acquisition function is optimized by first selecting from a pre-determined grid of points lying in $[0,1]^{23}$, distributed according to a Sobol sequence. Our baseline is a standard Gaussian process over this space that is agnostic to whether particular dimensions are irrelevant for a given point.

\paragraph{Results.}
\begin{figure}[t!]
	\centering
	\begin{subfigure}[]{0.45\textwidth}
		\includegraphics[width=\textwidth]{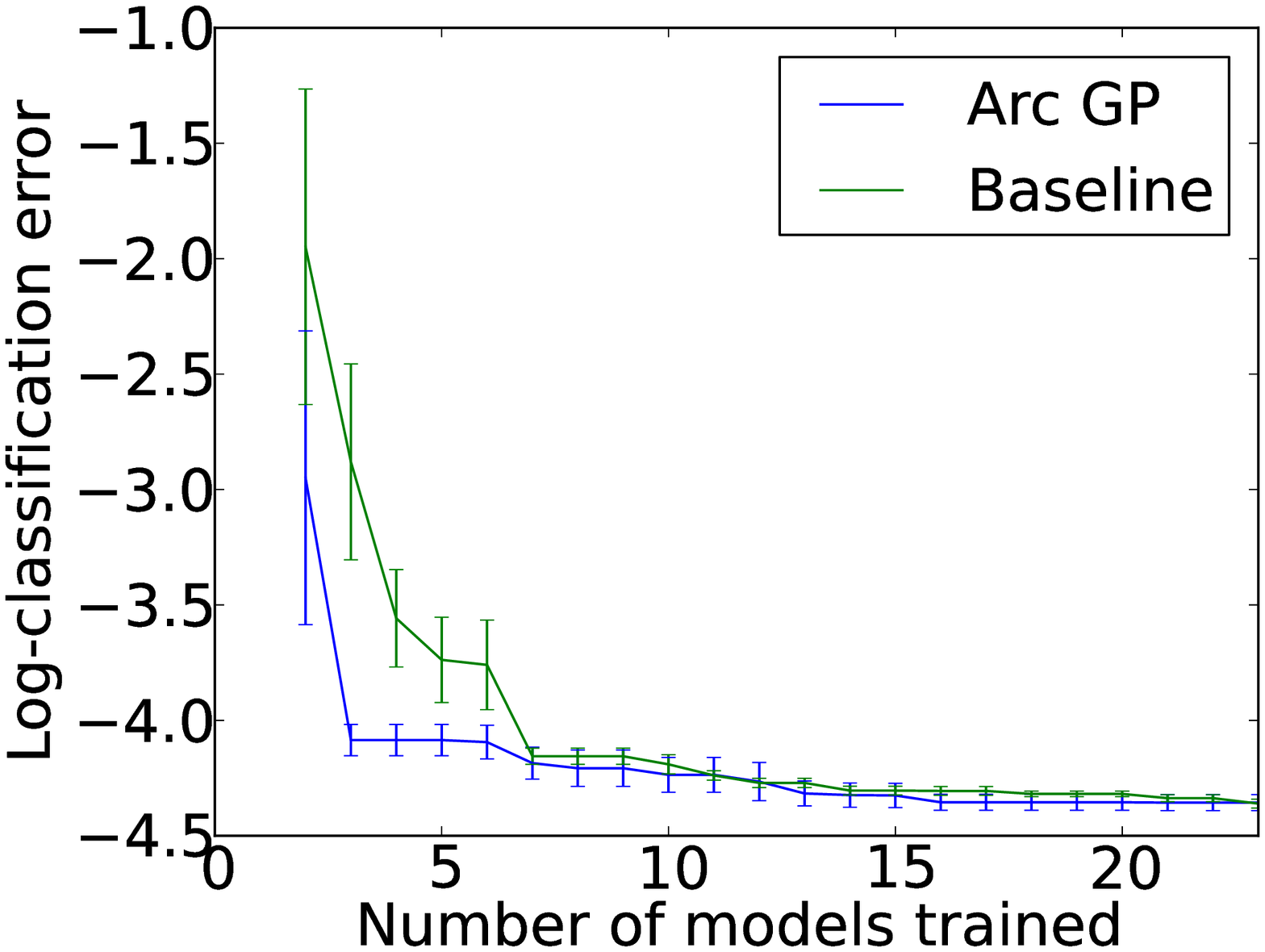}
		\caption{MNIST}
		\label{fig:mnist}
	\end{subfigure}\quad
	\begin{subfigure}[]{0.45\textwidth}
		\includegraphics[width=\textwidth]{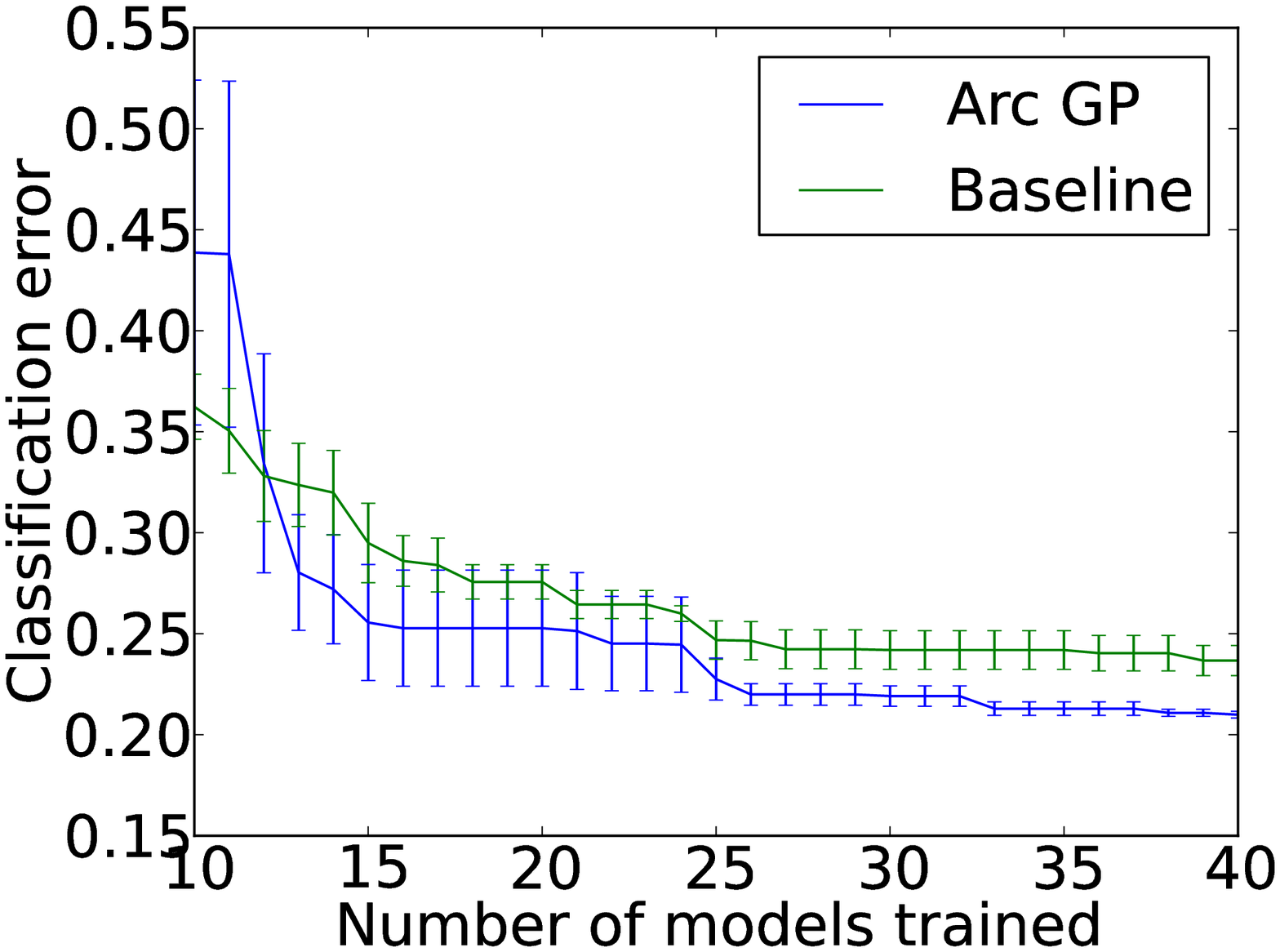}
		\caption{CIFAR-10}
		\label{fig:cifar10}
	\end{subfigure}
	\caption{Bayesian optimization results using the arc kernel.}
	\label{fig:arcbo}
%\vspace{-0.3cm}
\end{figure}

Figure~\ref{fig:arcbo} shows that on these datasets, using the arc kernel consistently reaches good solutions faster than the naive baseline, or it finds a better solution. In the case of MNIST, the best discovered model achieved $1.19\%$ test error using $50000$ training examples. By comparison,~\cite{wan2013regularization} achieved $1.28\%$ test error using a similar model and $60000$ training examples. Similarly, our best model for CIFAR-10 achieved $21.1\%$ test error using $45000$ training examples and $400$ features. For comparison, a support vector machine using $1600$ features with the same feature pipeline and $50000$ training examples achieves $22.1\%$ error.

\begin{figure}
	\includegraphics[width=0.45\textwidth]{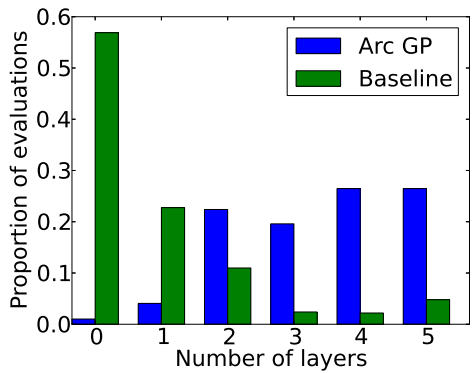}
	\caption{Relative fraction of neural net architectures searched on the CIFAR-10 dataset.}
	\label{fig:proparchs}
\end{figure}

Figure \ref{fig:proparchs} shows the proportion of function evaluations spent on each architecture size for the CIFAR-10 experiments. Interestingly, the baseline tends to favour smaller models while a \gp{} using the arc kernel distributes its efforts amongst deeper architectures that tend to yield better results.

\section{Conclusion}
%\vspace{-0.05in} 

We introduced the arc kernel for conditional parameter spaces that facilitates modelling the performance of deep neural network architectures by enabling the sharing of information across architectures where useful.
Empirical results show that this kernel improves \gp{} model quality and \gp{}-based Bayesian optimization results over several simpler baseline kernels. Allowing information to be shared across architectures improves the efficiency of Bayesian optimization and removes the need to manually search for good architectures. The resulting models perform favourably compared to established benchmarks by domain experts.

\section{Acknowledgements}
The authors would like to thank Ryan P. Adams for helpful discussions.

\bibliography{hierarchicalkernel}
\bibliographystyle{unsrt}

\end{document}